\documentclass[10pt,twocolumn,letterpaper]{article}

\usepackage{multirow}
\usepackage{pifont}
\usepackage[pagenumbers]{cvpr} 

\usepackage[table,dvipsnames]{xcolor}
\newcommand{\myparagraph}[1]{\vspace{2pt}\noindent{\bf #1}}
\usepackage{mathtools}

\usepackage{cuted}
\usepackage{capt-of}

\definecolor{cvprblue}{rgb}{0.21,0.49,0.74}
\usepackage[pagebackref,breaklinks,colorlinks,allcolors=cvprblue]{hyperref}

\title{DePT3R: Joint Dense Point Tracking and 3D Reconstruction of Dynamic Scenes in a Single Forward Pass}

\author{Vivek Alumootil \qquad Tuan-Anh Vu \\
University of California, Los Angeles
}

\begin{document}
\maketitle

\begin{strip}
    \centering
    \includegraphics[width=0.96\linewidth]{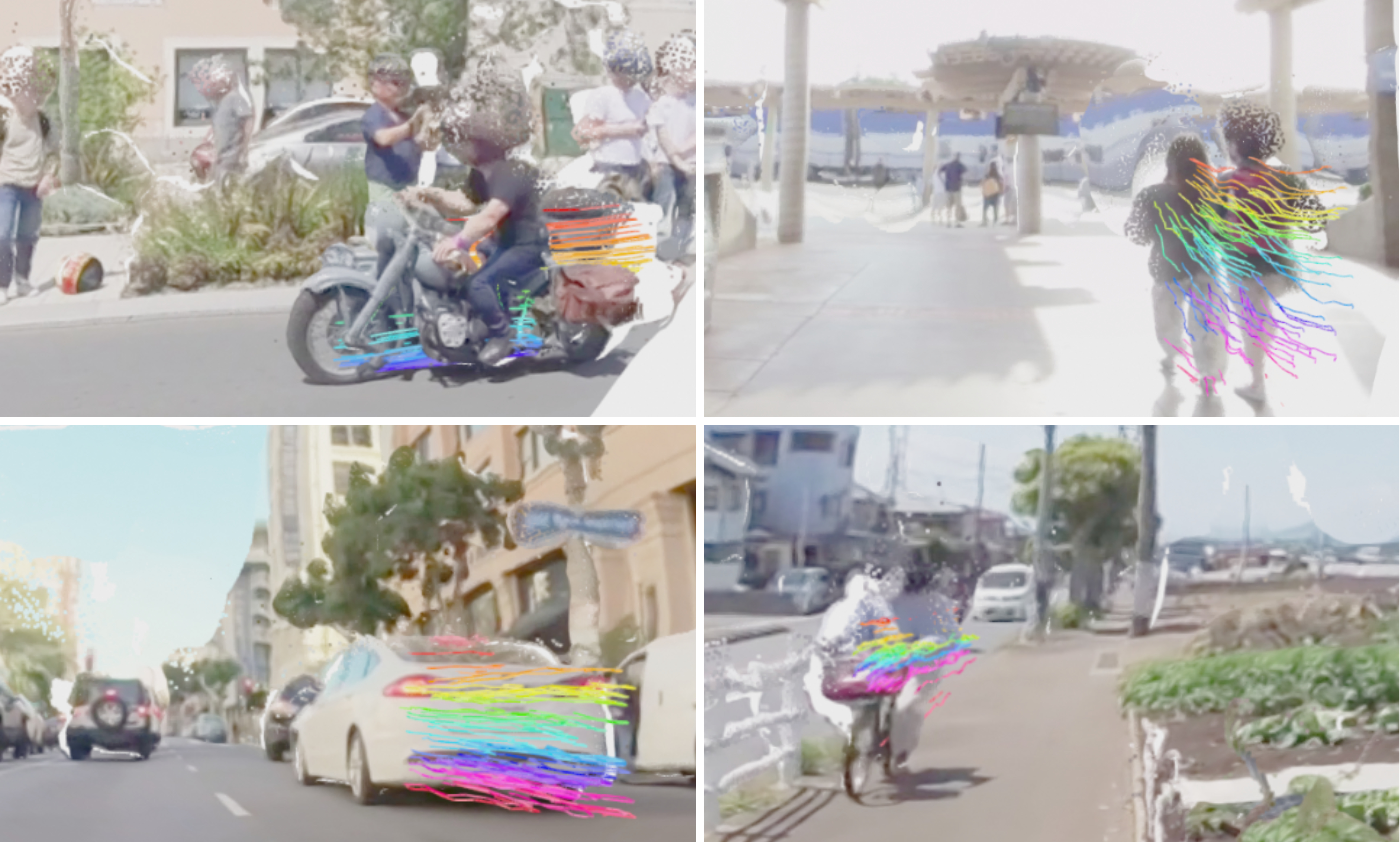}
    \vspace{-2mm}
    \captionof{figure}{\textbf{DePT3R} achieves \textit{robust} dense point tracking and reconstruction accuracy across unposed sequences while requiring \textit{less} memory usage, highlighting the effectiveness of our approach for \textit{long-range, dynamic} scenes.}
    \label{fig:teaser}
    \vspace{-2mm}
\end{strip}

\begin{abstract}
Current methods for dense 3D point tracking in dynamic scenes typically rely on pairwise processing, require known camera poses, or assume temporal ordering of input frames, thereby constraining their flexibility and applicability. Additionally, recent advances have successfully enabled efficient 3D reconstruction from large-scale, unposed image collections, underscoring opportunities for unified approaches to dynamic scene understanding. Motivated by this, we propose \textbf{DePT3R}, a novel framework that \textit{simultaneously performs} dense point tracking and 3D reconstruction of \textit{dynamic scenes} from multiple images in a single forward pass. This multi-task learning is achieved by extracting deep spatio-temporal features with a powerful backbone and regressing pixel-wise maps with dense prediction heads. Crucially, \textbf{DePT3R} operates \textit{without} requiring camera poses, substantially enhancing its adaptability and efficiency—especially important in dynamic environments with rapid changes. We validate \textbf{DePT3R} on several challenging benchmarks involving dynamic scenes, demonstrating \textit{strong performance} and significant improvements in \textit{memory efficiency} over existing state-of-the-art methods. Data and codes are available via the open repository: \url{https://github.com/StructuresComp/DePT3R}
\end{abstract}

\section{Introduction}
\label{sec:intro}

Understanding 3D scenes from images remains a foundational challenge in computer vision, with wide-ranging implications in autonomous navigation, augmented reality, and robotics. The ability to accurately track points and reconstruct 3D structures within complex and dynamic environments is vital for enabling intelligent systems to perceive and interact effectively with the real world~\cite{tavu2022rfnet4d,tavu2024rfnet4d++}. Traditional methods typically rely on extensive processing pipelines and auxiliary inputs, including depth maps and precise camera parameters, none of which are explicitly required for human visual perception. Moreover, these conventional approaches often fall short when handling dynamic scenes, which constitute a substantial portion of real-world environments, highlighting a significant gap between computational perception capabilities and human visual competence~\cite{song2022dynavins}.

Recent advances, notably DUSt3R~\cite{DUSt3R}, have significantly pushed the boundaries of pose-free 3D reconstruction. DUSt3R employs an extensively trained asymmetric transformer that implicitly aligns image features via cross-attention, enabling strong performance in static-scene reconstruction and downstream tasks such as point correspondence and scene flow estimation. While originally designed for static environments, DUSt3R also exhibits promising behavior on dynamic content, motivating further investigation into unified reconstruction and tracking. However, DUSt3R relies on pairwise image processing, which limits scalability and efficiency when applied to longer sequences and dynamic scenes. Subsequent work has therefore explored globally aggregated attention for multi-view reconstruction, achieving improved scalability and accuracy in static settings (e.g., Fast3R~\cite{yang2025fast3r}, VGGT~\cite{VGGT}).

Among learning-based global formulations, VGGT~\cite{VGGT} is a strong baseline that directly predicts camera parameters, pointmaps, depth, and 2D point tracks in a feed-forward manner, thereby avoiding additional optimization stages. Despite its strong reconstruction capabilities, VGGT exhibits two key limitations for dynamic scene understanding: \textit{(i)} it struggles under substantial non-rigid deformations, and \textit{(ii)} its memory footprint constrains dense tracking and limits efficiency when processing many frames jointly.

Motivated by these limitations, we introduce \textbf{DePT3R}, a feed-forward framework that \textit{jointly} performs dense point tracking and 3D reconstruction of dynamic scenes from unposed monocular image sequences via global image aggregation (Figure~\ref{fig:teaser}). Instead of chaining pairwise correspondences across time, DePT3R adopts a \textit{frame-to-query} formulation: given an observation time $t$ and a query time $q$, the model predicts per-frame geometry (pointmaps/depth) and a motion field that directly maps points from $t \rightarrow q$ in a single forward pass. This design enables long-range deformation reasoning without accumulating drift from frame-to-frame composition, while retaining the simplicity and efficiency of a feed-forward pipeline.

Empirically, we evaluate DePT3R on multiple dynamic-scene benchmarks using the metrics and protocols described in Section~\ref{sec:exp}. Under this evaluation, DePT3R improves 3D point tracking accuracy over prior learning-based baselines on PointOdyssey, DynamicReplica, and Panoptic Studio, and yields comparable reconstruction quality on PointOdyssey and TUM RGB-D (Table~\ref{tab:comp}). We further observe that models trained on short clips generalize to longer test sequences in our setting, and we report memory measurements indicating improved feasibility for dense tracking compared to query-based tracking in the same resolution regime (Table~\ref{tab:comp}).

In summary, we make the following contributions:
\begin{itemize}
    \item We introduce \textbf{DePT3R}, a framework that \textbf{jointly} performs dense point tracking and 3D reconstruction from unposed monocular image sequences \textbf{without} requiring auxiliary depth or external pose inputs.
    \item We propose a frame-to-query formulation that predicts a motion field mapping points from each observation time to a specified query time, enabling tracking \textbf{without} explicit frame-to-frame chaining.
    \item We extend a globally aggregated transformer backbone with a dedicated motion head, query conditioning, and an intrinsic embedding to incorporate camera intrinsics.
    \item We validate DePT3R using PointOdyssey, DynamicReplica, and Panoptic Studio for 3D point tracking. We use PointOdyssey and TUM RGB-D for reconstruction and provide memory measurements for dense tracking, compared with a query-based tracker operating at a similar resolution.
\end{itemize}

\begin{table*}[!th]
    \centering
    \caption{\textbf{Unposed 3D Reconstruction Methods Comparison.} We evaluate the key capabilities of prominent 3D reconstruction and point-tracking approaches, including support for dynamic scenes, 3D reconstruction, point tracking, and temporal attention. The results are tested on the Panoptic Studio dataset and the TUM RGB-D SLAM benchmark.  $\mathbf{\ast}$ Note that the VGGT method only has 2D point tracking. The best results are \textbf{bold}, and the second-best results are \underline{underline}.}
    \label{tab:comp}
    \vspace{-2mm}
    \setlength{\tabcolsep}{6pt}
    \renewcommand{\arraystretch}{1.2}
    \resizebox{0.94\linewidth}{!}{%
    \begin{tabular}{l|cccc|cc|cc}
    \toprule
    \multirow{2}{*}{\textbf{Method}} & \multirow{2}{*}{\textbf{\begin{tabular}[c]{@{}c@{}}Dynamic\\ Scene\end{tabular}}} & \multirow{2}{*}{\textbf{\begin{tabular}[c]{@{}c@{}}Unposed\end{tabular}}} & \multirow{2}{*}{\textbf{\begin{tabular}[c]{@{}c@{}}Dense Point\\ Tracking\end{tabular}}} & \multirow{2}{*}{\textbf{\begin{tabular}[c]{@{}c@{}}Temporal\\ Attention\end{tabular}}} & \multicolumn{2}{c|}{\textbf{Point Tracking}} & \multicolumn{2}{c}{\textbf{3D Reconstruction}} \\ \cmidrule(l){6-9} 
                                     &                                                                                   &                                                                                       &                                                                                    &                                                                                        & \textbf{APD} ($\uparrow$)         & \textbf{EPE} ($\downarrow$)        & \textbf{APD} ($\uparrow$)          & \textbf{EPE} ($\downarrow$)         \\ \midrule
    DUSt3R~\cite{DUSt3R}                           & -                                                                                 & \ding{51}                                                             & -                                                                                  & -                                                                                      & -                     & -                    & 72.27                  & 0.29                  \\
    MonST3R~\cite{zhang2025monst3r}                          & \ding{51}                                                         & \ding{51}                                                             & \ding{51}                                                                                  & -                                                                                      & 51.32                 & 0.46                 & 61.38                  & 0.36                  \\
    MASt3R~\cite{mast3r}                           & -                                                                                 & \ding{51}                                                             & \ding{51}                                                                                  & -                                                                                      & -                     & -                    & 66.22                  & 0.55                  \\
    St4RTrack~\cite{st4rtrack2025}                        & \ding{51}                                                         & \ding{51}                                                             & \ding{51}                                                          & -                                                                                      & \underline{69.67}                 & \underline{0.26}                 & 83.42                  & 0.19                  \\
    SpatialTracker~\cite{xiao2024spatialtracker}                   & \ding{51}                                                                                 & -                                                                                     & -                                                          & \ding{51}                                                                                      & 62.59                 & 0.31                 & -                      & -                     \\
    VGGT~\cite{VGGT}$\ast$                       & \ding{51}                                                                                 & \ding{51}                                                             & -                                                                                  & \ding{51}                                                              & -                     & -                    & \underline{89.87}                      & \textbf{0.09}                     \\ \midrule
    DePT3R (Ours)                    & \ding{51}                                                         & \ding{51}                                                             & \ding{51}                                                          & \ding{51}                                                              & \textbf{89.36}                   & \textbf{0.10}                  & \textbf{92.22}                    & \underline{0.10}                   \\ \bottomrule
    \end{tabular}
    }
    \vspace{-2mm}
\end{table*}

\vspace{-1mm}
\section{Related Works}
\label{sec:rw}

\subsection{Tabula Rasa 3D Reconstruction}
\label{subsec:3drec}

Tabula rasa methods aim to reconstruct 3D scenes solely from raw input observations, without relying on prior scene knowledge. Early techniques, such as Structure from Motion (SfM), relied heavily on matching handcrafted features across multiple source images to estimate camera poses and reconstruct scene structure~\cite{hartley2003multiple}. The advent of Neural Radiance Fields (NeRF)~\cite{martin2021nerf} initiated a paradigm shift by formulating reconstruction as an optimization-based novel-view synthesis problem, where scenes are represented as continuous neural fields. InstantNGP~\cite{muller2022instant} accelerated this optimization by coupling a compact neural network with a multi-resolution hash table. Further improvements in rendering speed were achieved by 3D Gaussian Splatting~\cite{3dgs}, which explicitly represents scenes as collections of anisotropic Gaussian functions, efficiently rasterized for real-time rendering. Subsequently, Mip-Splatting~\cite{Yu2024MipSplatting} reduced rendering aliasing by constraining Gaussian sizes. Despite impressive results in view synthesis, these approaches necessitate per-scene optimization and cannot leverage learned data priors.

\subsection{Learning-based 3D Reconstruction}
\label{subsec:learning3drec}

Deep learning has driven rapid progress in 3D reconstruction. Early methods employed multi-stage pipelines with learned feature extractors or end-to-end models, focusing on cost volumes and correlation operations~\cite{wang2024deep}. More recently, transformer-based architectures have gained prominence, mitigating the limitations of CNNs, restricted receptive fields, and weak long-range dependency modeling~\cite{zhu2021multi}.

One notable breakthrough was DUSt3R~\cite{DUSt3R}, which enabled 3D reconstruction from unposed image pairs using asymmetric transformers linked via cross-attention. The predictions from individual image pairs were subsequently globally optimized in a standard coordinate system. Subsequent developments included symmetric architectures and global attention layers, which significantly improved memory efficiency for processing larger image sets~\cite{yang2025fast3r}.

MASt3R~\cite{mast3r} built on DUSt3R's architecture by adding additional point-map predictions, allowing for more accurate pixel matching. Furthermore, recent methods have combined DUSt3R-inspired reconstruction strategies with efficient Gaussian splatting-based rendering, facilitating rapid, feed-forward novel-view synthesis~\cite{smart2024splatt3r,fan2024instantsplat,chen2025dense,chen2024pref3r}.

Adapting DUSt3R for dynamic scenes, MonST3R~\cite{zhang2025monst3r}  demonstrated its effectiveness through targeted fine-tuning. CUT3R~\cite{wang2025cut3r} further developed this concept by introducing an online, recurrent reconstruction framework that incrementally processes images and updates scene reconstructions. Recent advances in learning-based 3D reconstruction have culminated in VGGT~\cite{VGGT}, which serves as a robust foundation for 3D reconstruction. This method employs a large transformer-based architecture to predict essential 3D attributes, including camera intrinsics, point maps, depth maps, and point tracks.
By directly predicting these attributes, VGGT eliminates the need for post-processing, achieving state-of-the-art results in 3D point and camera pose reconstruction. Despite these advancements, challenges remain due to substantial camera and object motion, underscoring the need for further exploration in this area.

\subsection{2D Point Tracking}
\label{subsec:2dpoint}

Optical flow estimation, traditionally framed as an energy minimization problem, was transformed by FlowNet~\cite{dosovitskiy2015flownet}, a pioneering deep learning-based solution that leverages convolutional neural networks. RAFT~\cite{teed2020raft} introduced an iterative refinement strategy, querying a 4D cost volume for flow updates. Flowformer~\cite{huang2022flowformer} improved upon this by transforming the 4D cost volume into tokenized representations processed by transformer layers.

Beyond pairwise flow, recent research has advanced pixel tracking across multiple frames. Particle Video (PIPs)~\cite{harley2022particle} revisited multi-frame point trajectory estimation, leveraging temporal priors to improve occlusion handling, but its eight-frame context window limited its applicability. TAP-Vid~\cite{doersch2022tap} reformulated the tracking problem, established a benchmark, and introduced TAP-Net as a simple baseline. CoTracker~\cite{karaev2024cotracker} recognized statistical interdependencies among trajectories, utilizing transformers to estimate large sets of point tracks jointly.

\subsection{3D Point Tracking}
\label{subsec:3dpoint}

Estimation of 3D point motion has also advanced significantly. RAFT-3D~\cite{teed2021raft} employed rigid motion embeddings to estimate scene flow between RGBD pairs. Omnimotion~\cite{wang2023tracking} unified 2D and 3D tracking with a quasi-3D canonical representation but still required per-scene optimization. Introduced by Xiao et al.~\cite{xiao2024spatialtracker}, SpatialTracker enables feed-forward tracking by creating a triplane scene encoding with depth estimation. The core mechanism is a transformer that iteratively refines query point trajectories, producing updated paths using the prior trajectory, the query point's features, and the features of nearby points. DELTA~\cite{ngo2025delta} efficiently computed dense 3D trajectories, while TAPIP3D~\cite{zhang2025tapip3d} leveraged depth to elevate image features into a global coordinate system for enhanced performance. Building on the DUSt3R architecture, St4RTrack~\cite{st4rtrack2025} adapted it for 3D point tracking by incorporating an additional prediction head and chaining motion predictions for longer trajectories. However, its pairwise processing limits the exploitation of temporal attention, thereby restricting its ability to handle significant camera motion effectively. POMATO~\cite{zhang2025pomato} addressed these limitations by introducing a temporal attention module, eliminating scale normalization, and enhancing inter-frame interactions. Nonetheless, its fundamentally pairwise processing strategy continued to restrict fully global interactions, indicating room for further advancement. Stereo4D ~\cite{jin2024stereo4d} attached a temporal motion head to the DUSt3R architecture used to estimate the position of scene points at any query time between the time of the two input frames and introduced a large dataset of internet videos with noisy point trajectories to train the motion head.

\begin{figure*}[t]
    \centering
    \includegraphics[width=0.98\linewidth]{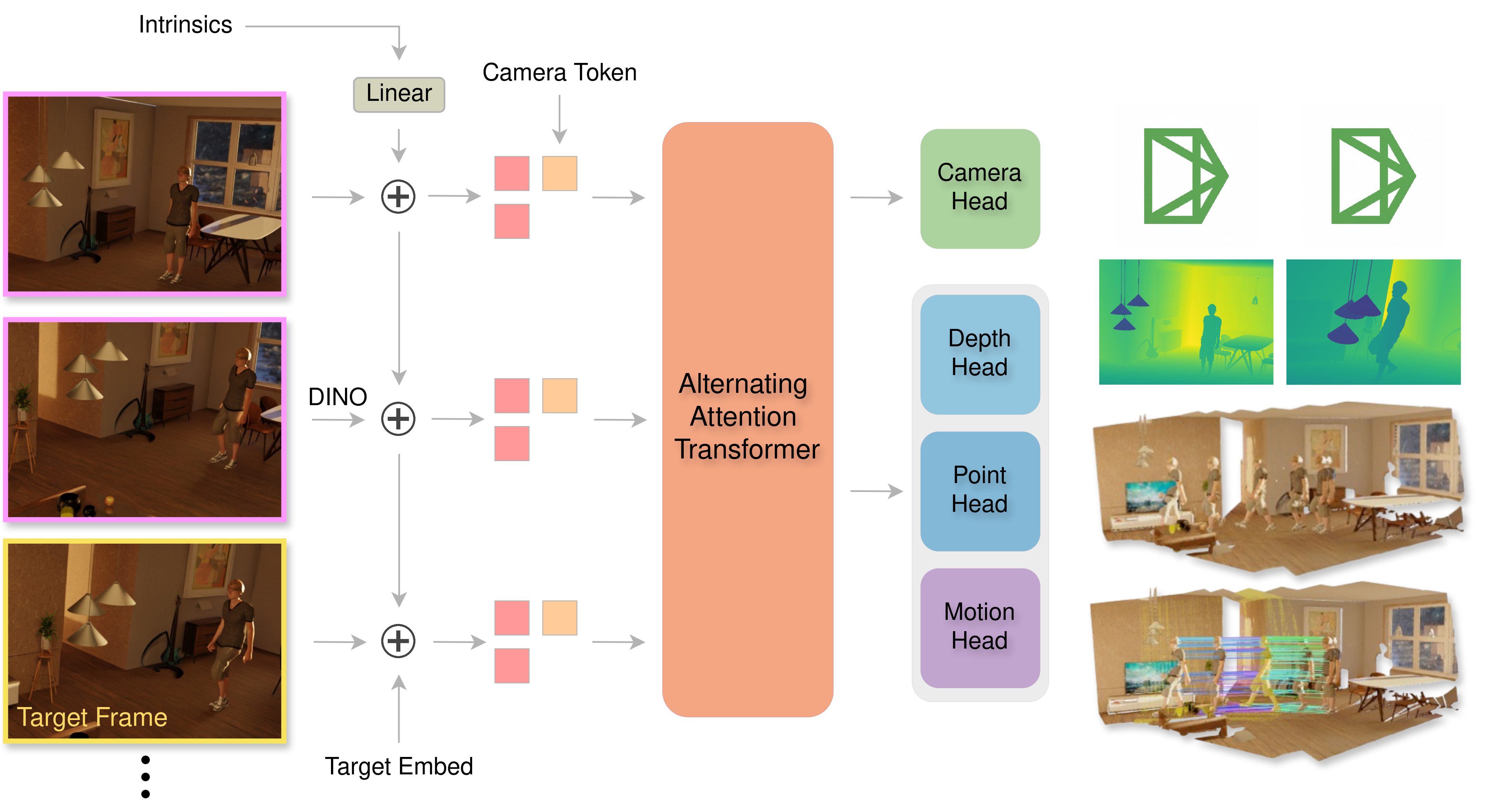}
    \caption{\textbf{Our proposed \textbf{DePT3R} framework.} Our model first tokenizes each input frame with DINOv2 and augments every token with a global intrinsic embedding. A learnable query embedding is added to the tokens corresponding to the query frame. The alternating frame-wise and global self-attention blocks process the tokens. A dedicated camera head predicts both intrinsics and extrinsics, while DPT heads produce point maps, depth maps, and motion maps (3D point tracks).}
    \label{fig:framework}
\end{figure*}

\section{Proposed Method}
\label{sec:method}

\textbf{DePT3R} jointly produces dense point tracks and reconstructs dynamic scenes from a sequence of RGB inputs with a single forward pass. An overview is shown in Figure~\ref{fig:framework}.

\subsection{Preliminary}
\label{subsec:pre}

\myparagraph{Pointmap Representation.} We adopt the time-dependent pointmap representation introduced by St4RTrack~\cite{st4rtrack2025}. This representation assumes that every pixel in an image corresponds to a 3D point and maps each pixel to a specific 3D position. A pointmap $\prescript{c}{}{\mathbf{X}}^a_t \in \mathbb{R}^{H \times W \times 3}$ encodes the 3D positions of scene points visible in frame $a$, at the time of frame $t$, expressed within frame $c$'s coordinate system. Specifically, $\prescript{c}{}{\mathbf{X}}^a_t(i, j)$ denotes the 3D position, at the time of frame $t$, of the scene point corresponding to pixel $(i,j)$ in frame $a$, with this position given in frame $c$'s coordinate system. To illustrate, $\prescript{c}{}{\mathbf{X}}^a_1$ captures the 3D positions of scene points visible in frame $a$ at the time of frame $1$, while $\prescript{c}{}{\mathbf{X}}^a_2$ captures the positions of these same points at the time of frame $2$.

In a static scene, the time subscript can be omitted due to the absence of point motion. As noted by DUSt3R \cite{DUSt3R}, we can obtain $\prescript{a}{}{\mathbf{X}}^a$ by unprojecting the pixels of frame $a$ into 3D space with the intrinsic matrix $K_a$ and depth map $D$:
$$\prescript{a}{}{\mathbf{X}}^a(i, j) = K_a^{-1}[iD(i, j), jD(i, j), D(i, j)]^T,$$

Subsequently, $\prescript{c}{}{\mathbf{X}}^a$ can be readily obtained by transforming these 3D positions into frame $c's$ coordinate system using the camera poses of frames $a$ and $c$.

Conversely, in a dynamic scene, most point maps cannot be described solely by simple geometric transforms, as they inherently encode scene motion. Note that by leveraging $\prescript{c}{}{\mathbf{X}}^a_{t_1}$ and $\prescript{c}{}{\mathbf{X}}^a_{t_2}$, we can effectively analyze the motion of the scene points visible in frame $a$ from time of frame $t_1$ to the time of frame $t_2$.

\myparagraph{Feature Backbone.} We leverage Visual Grounded Geometry Transformer (VGGT)~\cite{VGGT}, a robust architecture for unposed 3D reconstruction trained on extensive static and dynamic datasets, as our core backbone.

In the VGGT method, each input image is first decomposed into image tokens via DINOv2 ~\cite{oquab2023dinov2}. These tokens, along with a learnable camera token and four learnable register tokens per image, are then input to a global aggregator module. Since predictions are made in the initial frame's coordinate system, the camera and register tokens for the first image differ from those for subsequent frames, enabling the network to discriminate among frames. The aggregator module consists of alternating local and global attention layers: local layers facilitate interactions within a single image, while global layers promote interactions across images.

Output tokens from the aggregator are processed through dedicated Dense Prediction Transformer (DPT) heads ~\cite{ranftl2021vision} —specifically, a pointmap DPT head and a depth DPT head—to regress corresponding pointmaps and depth maps. These heads also output aleatoric uncertainty maps. Camera tokens are further processed through four self-attention layers followed by a linear projection to predict normalized camera extrinsics and intrinsics.

\subsection{Point Tracking and Dynamic Reconstruction}
\label{subsec:proposed}

Our method takes a sequence of $N$ frames, $(I_{t})^{N}_{t=1}$, and a query time $q$, and jointly reconstructs the scene and tracks observable points in 3D. To this end, for each frame $I_t$, we predict two pixel-wise maps expressed in the coordinate system of the first frame $I_1 : \prescript{1}{}{\mathbf{\hat{X}}}^t_{t}$, capturing point positions at the observation time, and $\prescript{1}{}{\mathbf{\hat{M}}}^t_{q} = \prescript{1}{}{\mathbf{\hat{X}}}^t_{q}-\prescript{1}{}{\mathbf{\hat{X}}}^t_{t}$, capturing the motion of points from time of frame $t$ to the time of query frame $q$. By jointly estimating these maps, our method concurrently reconstructs the 3D structure and tracks point motion to the query frame. Notably, our tracking approach establishes direct correspondences between each frame and the query frame, rather than using pairwise frame-to-frame tracking, enabling efficient capture of overall motion and deformation over extended intervals.

We integrate an additional DPT head—the \emph{motion head}—to the VGGT network architecture to regress the motion map $\prescript{1}{}{\mathbf{M}}^t_{q}$. This structural integration preserves VGGT's pretrained capabilities while enabling motion estimation. Despite sharing identical inputs, the reconstruction and motion heads produce temporally distinct pointmaps. Prior works ~\cite{st4rtrack2025,liang2025zero} proved that shared image tokens can effectively encode both reconstruction and motion information, thus permitting accurate motion prediction without degrading reconstruction quality.

\subsection{Intrinsic Embedding}
Previous work ~\cite{ye2025no} has identified the importance of providing intrinsic information for accurate 3D reconstruction. Since camera intrinsic parameters are often readily available through calibration, we propose incorporating an intrinsic embedding to enable the model to leverage these crucial parameters directly. To implement the intrinsic embedding, we concatenate a subset of the normalized camera intrinsic parameters, specifically the focal lengths $f_x$ and $f_y$ and the principal point's y-coordinate $p_y$, and pass this vector through a linear layer. Note that the principal point's x-coordinate $p_x$ is excluded from this embedding because all input images are scaled to a fixed width of 514 pixels ~\cite{VGGT}. The resulting feature is then added to all image tokens, providing the model with a direct representation of the camera's intrinsic properties.

\subsection{Loss Function}
\label{subsec:loss}

Following VGGT, our multi-task loss is defined as:
\begin{equation}
L = \lambda_{c}L_{\mathrm{camera}} + \lambda_{d}L_{\mathrm{depth}} + \lambda_{p}L_{\mathrm{point}} + \lambda_{m}L_{\mathrm{motion}}
\end{equation}

The camera loss $L_{\mathrm{camera}}$ is computed using Huber loss:
\begin{equation}
L_{\mathrm{camera}} = \sum_{i=1}^{N} \lVert \hat{\mathbf{g}}_i - \mathbf{g}_i \rVert_{\epsilon}
\end{equation}
where $\mathbf{\hat{g}}_i$ denotes the ground-truth camera parameters and $\mathbf{g}_i$ the ground truth.

The depth loss ($L_{\mathrm{depth}}$) is a combination of three key components: regression loss ($L_{\mathrm{reg}}$), confidence loss ($L_{\mathrm{conf}}$), and gradient loss ($L_{\mathrm{grad}}$).
\begin{equation}
L_{\mathrm{depth}}= L_{\text{reg}} + L_{\text{conf}} + L_{\text{grad}}
\end{equation}

The regression loss ($L_{\mathrm{reg}}$) quantifies the average error between predicted ($\hat{D}_i$) and ground-truth depths ($D_i$) for all supervised pixels:
\begin{equation}
L_{\mathrm{reg}} = \frac{1}{N_{\mathrm{points}}} \sum_{i=1}^{N} \lVert \hat{D}_i-D_i\rVert
\end{equation}

The confidence loss ($L_{\mathrm{conf}}$) weights these errors using a predicted uncertainty map ($\Sigma_i^{D}$):
\begin{equation}
L_{\mathrm{conf}} = \frac{1}{N_{\mathrm{points}}} \sum_{i=1}^{N} \left (\lVert \Sigma_i^{D} \odot (\hat{D}_i-D_i)\rVert - \alpha \log({\Sigma_i^{D}}) \right)
\end{equation}

Finally, the gradient loss ($L_{\mathrm{grad}}$) promotes smoothness in the predicted depth map by comparing the gradients of the predicted and GT depths, weighted by uncertainty:
\begin{equation}
L_{\mathrm{grad}} = \frac{1}{N_{\mathrm{points}}} \sum_{i=1}^{N} \lVert \Sigma_i^{D} \odot (\nabla \hat{D}_i- \nabla D_i)\rVert
\end{equation}

We employ a similar approach for both point loss and motion loss. However, the motion loss does not include the gradient-based loss.

\begin{table*}[t]
\centering
\caption{\textbf{World Coordinate 3D Point Tracking.} Comparison of our \textbf{DePT3R} method with existing point tracking methods on PointOdyssey, DynamicReplica, and Panoptic Studio datasets. We report the APD and EPE metrics after global median scaling. The best results are \textbf{bold}, and the second-best results are \underline{underline}.}
\label{tab:tracking}
\vspace{-2mm}
\setlength{\tabcolsep}{5pt}
\renewcommand{\arraystretch}{1.2}
\resizebox{0.99\linewidth}{!}{%
\begin{tabular}{l|ccc|ccc}
\toprule
\multirow{2.2}{*}{\textbf{Method}} & \multicolumn{3}{c|}{\textbf{APD} ($\uparrow$)} & \multicolumn{3}{c}{\textbf{EPE} ($\downarrow$)} \\
\cmidrule(lr){2-4} \cmidrule(lr){5-7}
& \textbf{PointOdyssey} & \textbf{DynamicReplica} & \textbf{Panoptic Studio} & \textbf{PointOdyssey} & \textbf{DynamicReplica} & \textbf{Panoptic Studio} \\
\midrule \midrule
SpatialTracker~\cite{xiao2024spatialtracker} & 38.54 & 54.85 & 62.59 & 0.7499 & 0.9274 & 0.3094 \\
MonST3R~\cite{zhang2025monst3r} & 33.47 & 58.06 & 51.32 & 0.9021 & 0.4387 & 0.4568 \\
St4RTrack~\cite{st4rtrack2025} & \underline{67.95} & \underline{73.74} & \underline{69.67} & \underline{0.3140} & \underline{0.2682} & \underline{0.2637} \\
\midrule
DePT3R (Ours) & \textbf{91.33} & \textbf{91.12} & \textbf{89.36}& \textbf{0.0949} & \textbf{0.0925} & \textbf{0.1046} \\
\bottomrule
\end{tabular}
}
\vspace{-2mm}
\end{table*}

\section{Experiments}
\label{sec:exp}

\subsection{Experimental Setups}
\label{subsec:impl}

We designed our experimental setup to evaluate the scalability and versatility of \textbf{DePT3R} across diverse environments, utilizing large datasets and comprehensive metrics that assess both tracking accuracy and reconstruction quality.

\myparagraph{Datasets.} 
Our training was conducted on five synthetic datasets. PointOdyssey (PO)~\cite{zheng2023pointodyssey}, DynamicReplica (DR)~\cite{karaev2023dynamicstereo}, and Kubric Movi-F~\cite{greff2022kubric} offer camera and scene motion, as well as ground truth mesh vertex trajectories, which we use for sparse point tracking supervision during training. Virtual KITTI 2 ~\cite{cabon2020virtual} and TartanAir ~\cite{wang2020tartanair} do not include point trajectories; therefore, we only use them to supervise camera poses, depth maps, and point maps. 

We evaluate our method on PointOdyssey and DynamicReplica, as well as Panoptic Studio (PS)~\cite{joo2015panoptic} and the TUM RGB-D SLAM Benchmark~\cite{sturm2012benchmark}. We also include a qualitative evaluation on the Stereo4D dataset ~\cite{jin2024stereo4d}. We do not evaluate on the Aria Digital Twin Benchmark~\cite{koppula2024tapvid3d} because the VGGT method \textbf{was trained} on it. 

\myparagraph{Baselines.} We selected recent state-of-the-art (SOTA) methods for comparison, including three point tracking methods (SpatialTracker~\cite{xiao2024spatialtracker}, MonST3R~\cite{zhang2025monst3r}, and St4RTrack~\cite{st4rtrack2025}) and five 3D reconstruction methods (DUSt3R~\cite{DUSt3R}, MASt3R~\cite{mast3r}, MonST3R~\cite{zhang2025monst3r}, St4RTrack~\cite{st4rtrack2025}, and VGGT~\cite{VGGT}).

\myparagraph{Metrics.} We use two metrics to evaluate the efficacy and quality of our method: \textbf{APD} and \textbf{EPE} metrics. 

The \textbf{APD metric} was originally proposed by TAPVid-3D~\cite{koppula2024tapvid3d} for tracking in 3D. Following the methodology of St4RTrack~\cite{st4rtrack2025}, we calculate the APD as the average percentage of predicted points whose error falls below a set of thresholds. Specifically, let $\hat{P_t^i}$ denote the prediction for the $i$th point at time $t$, let $P_t^i$ denote the ground truth position. The \textbf{APD metric} is calculated as below:
\begin{equation}
\mathrm{APD}_{\mathrm{3D}} = \frac{1}{4N_{\mathrm{points}} }\sum_{d \in \delta_{\mathrm{3D}}} \sum_{i, t} \mathbf{1}(\lVert \hat{P_t^i} - sP_t^i   \rVert < d)
\end{equation}
where $\delta_{3D} = \{\mathrm{0.1m, 0.3m, 0.5m, 1.0m}\}$. Since predictions are made in the coordinate system of the first frame's camera, we align the estimated 3D positions with the ground truth by scaling them by the median ratio of the norms of the estimated and ground-truth positions. That is, we multiply the predicted coordinates by the $s$ factor: 
\begin{equation}
s = \frac{\mathrm{median}_{i,t}(\lVert P_t^i \rVert)}{\mathrm{median}_{i,t}(\lVert \hat{P_t^i} \rVert)}
\end{equation}

The \textbf{EPE metric} is defined as the average Euclidean distance between the ground truth and scaled predicted positions for all points:
\begin{equation}
\mathrm{EPE} = \frac{1}{N_{\mathrm{points}}} \sum_{i, t} \lVert s\hat{P_t^i} - P_t^i \rVert 
\end{equation}

\myparagraph{Implementation Details.}
To generate the ground-truth motion maps, we project the visible 3D mesh vertices for each frame onto the image plane using the known camera extrinsics and intrinsics, then round to the nearest integer pixel coordinate. Mesh vertices are excluded if they are not visible in the query frame. 

To effectively utilize VGGT's extensive pretraining knowledge, we initialize the weights of the alternating attention transformer, point head, and depth head with those of VGGT, and the motion head with those of the VGGT pointmap head. 

During training, sequences of 2 to 10 frames are randomly selected from all available sequences. For Kubric Movi-F, we use a stride of $1$; for all other datasets, we use a stride ranging from $1$ to $4$. TartanAir and Virtual KITTI 2 are sampled with half the frequency of the different datasets. 

We observe that training with the final tracking objective from the start, using model weights initialized with VGGT's weights, degrades performance. Thus, we perform training in two phases. In the first phase, we train \textbf{DePT3R} using only the camera, depthmap, and pointmap losses. The intrinsic embedding is used in the first phase, but the query embedding is not. In the second phase, we integrate the query embedding and motion head, along with the tracking loss. To improve training stability and speed, we do not use a confidence-based loss in the second phase.

We employ common data augmentation strategies, such as color jittering, random aspect ratios, and random center crops ~\cite{DUSt3R}, applied uniformly to each image in the sequence. We use the Adam optimizer with a warmup phase, followed by cosine scheduling. Gradient checkpointing and bfloat16 precision are utilized to enhance efficiency and reduce GPU memory usage. In the first phase, we train on all datasets, using a learning rate of $5 \times 10^{-5}$ for the intrinsic embedding and a learning rate of $5 \times 10^{-6}$ for the other weights. In the second phase, we train only on PointOdyssey, DynamicReplica, and Kubric, using a learning rate of $1 \times 10^{-5}$ for the motion head and the query embedding, and a learning rate of $1 \times 10^{-6}$ for all other weights. Both training phases are run on a single NVIDIA RTX 6000 and take approximately 3 days. We use PyTorch 2.3.1 as our machine learning framework. Experiments are performed on a machine equipped with an AMD Ryzen Threadripper PRO 7975WX 32-core CPU and Ubuntu 22.04 installed.

\begin{table}[!t]
    \centering
    \caption{\textbf{World Coordinate 3D Reconstruction.} Comparison of our \textbf{DePT3R} method with existing reconstruction methods on PO and TUM datasets. We report the APD and EPE metrics after global median scaling. The best results are \textbf{bold}, and the second-best results are \underline{underline}. }
    \label{tab:recon}
    \vspace{-2mm}
    \setlength{\tabcolsep}{5pt}
    \renewcommand{\arraystretch}{1.3}
    \resizebox{0.99\linewidth}{!}{%
    \begin{tabular}{l|cc|cc} 
    \toprule
    \multirow{2}{*}{\textbf{Method}} & \multicolumn{2}{c|}{\textbf{PointOdyssey}} & \multicolumn{2}{c}{\textbf{TUM RGB-D SLAM}} \\
    \cmidrule(lr){2-3} \cmidrule(lr){4-5} 
    & \textbf{APD} ($\uparrow$) & \textbf{EPE} ($\downarrow$) & \textbf{APD} ($\uparrow$) & \textbf{EPE} ($\downarrow$) \\
    \midrule \midrule
    DUSt3R~\cite{DUSt3R} & 45.79 & 0.6386 & 72.27 & 0.2891 \\
    MASt3R~\cite{mast3r} & 56.90 & 0.4644 & 66.22 & 0.5510 \\
    MonST3R~\cite{zhang2025monst3r} & 68.25 & 0.3044 & 61.38 & 0.3646 \\
    St4RTrack~\cite{st4rtrack2025} & 78.73 & 0.2406 & 83.42 & 0.1854 \\
    VGGT~\cite{VGGT} & \underline{97.69} & \underline{0.0514} & \underline{89.87} & \textbf{0.0930} \\
    \midrule
    DePT3R (Ours) & \textbf{98.01} & \textbf{0.0406} & \textbf{92.22} & \underline{0.0968} \\
    \bottomrule
    \end{tabular}
    }
    \vspace{-2mm}
\end{table}

\begin{figure*}[t]
    \centering
    \includegraphics[width=0.99\linewidth]{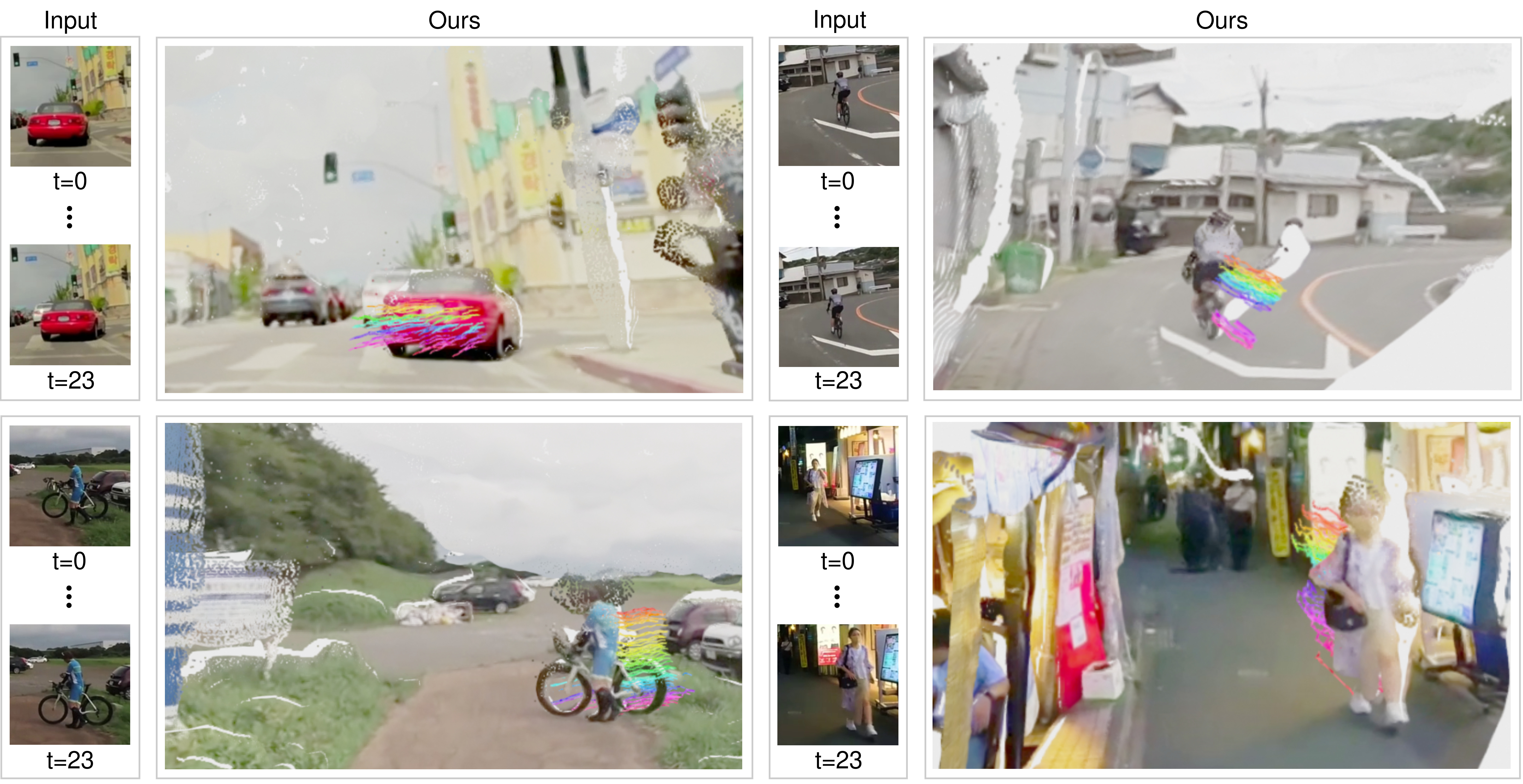}
    \caption{\textbf{3D Reconstruction and Point Tracking on the Stereo4D Dataset.} Despite supervising point tracking on a small collection of unrealistic datasets with mostly minimal scene motion, our method exhibits strong generalization to real-world scenes.}
    \label{fig:vis}
\end{figure*}

\subsection{Experimental Results}

\myparagraph{3D Point Tracking.} We first evaluate the 3D point tracking task across three datasets: the test sets of the synthetic datasets PointOdyssey and DynamicReplica, which feature significant camera and scene motion, and PanopticStudio, a real-world dataset with no camera motion. For evaluation, we followed the same setup of St4RTrack~\cite{st4rtrack2025} and selected 64 consecutive frames from 50 randomly selected sequences in each dataset. To compare with previous work, we restrict our evaluation to the trajectories of the points from the first frame, even though our method can track all points. Following St4RTrack, we exclude PointOdyssey sequences that contain fog or are generated in the Kubric style. 

As shown in Table~\ref{tab:tracking}, our method significantly outperforms all baselines across all datasets. Remarkably, even though our model was trained only on sequences of at most 10 images, it achieves impressive performance on much longer sequences. Moreover, unlike other approaches that require windowing or global optimization to handle extended sequences due to memory or architectural constraints, our method can generate predictions for a large number of frames in a single forward pass. 

\myparagraph{3D Reconstruction.} Following St4RTrack~\cite{st4rtrack2025}, we diverge from previous work, directly evaluating the accuracy of our estimated 3D reconstruction with the APD and EPE metrics. We use the TUM RGB-D Benchmark, a real-world dataset with significant camera and scene motion, and the PointOdyssey test set. We report metrics on 50 randomly chosen sequences of 64 consecutive frames. For the TUM RGB-D Benchmark, we filter out points with depths between 0.1 and 5 meters, as the depth camera's accuracy degrades at long ranges. On the TUM RGB-D Benchmark (Table~\ref{tab:recon}), \textbf{DePT3R} achieves an APD of 92.22 and an EPE of 0.0968, competitive with VGGT and significantly better than the other baselines. On PointOdyssey, \textbf{DePT3R} achieves an EPE of 0.0406 and an APD of 98.01, noticeably better than all of the baselines. These results demonstrate that \textbf{DePT3R} achieves state-of-the-art 3D reconstruction accuracy on both synthetic and real-world data, including out-of-distribution scenarios, using synthetic training data.

We showed visualization of our method for both 3D point tracking and 3D reconstruction in Figure~\ref{fig:vis}. Our method demonstrates strong generalization to large, realistic datasets with significant scene motion, despite being trained on a small collection of unrealistic datasets with minimal scene motion.

\begin{figure*}[t]
    \centering
    \includegraphics[width=0.99\linewidth]{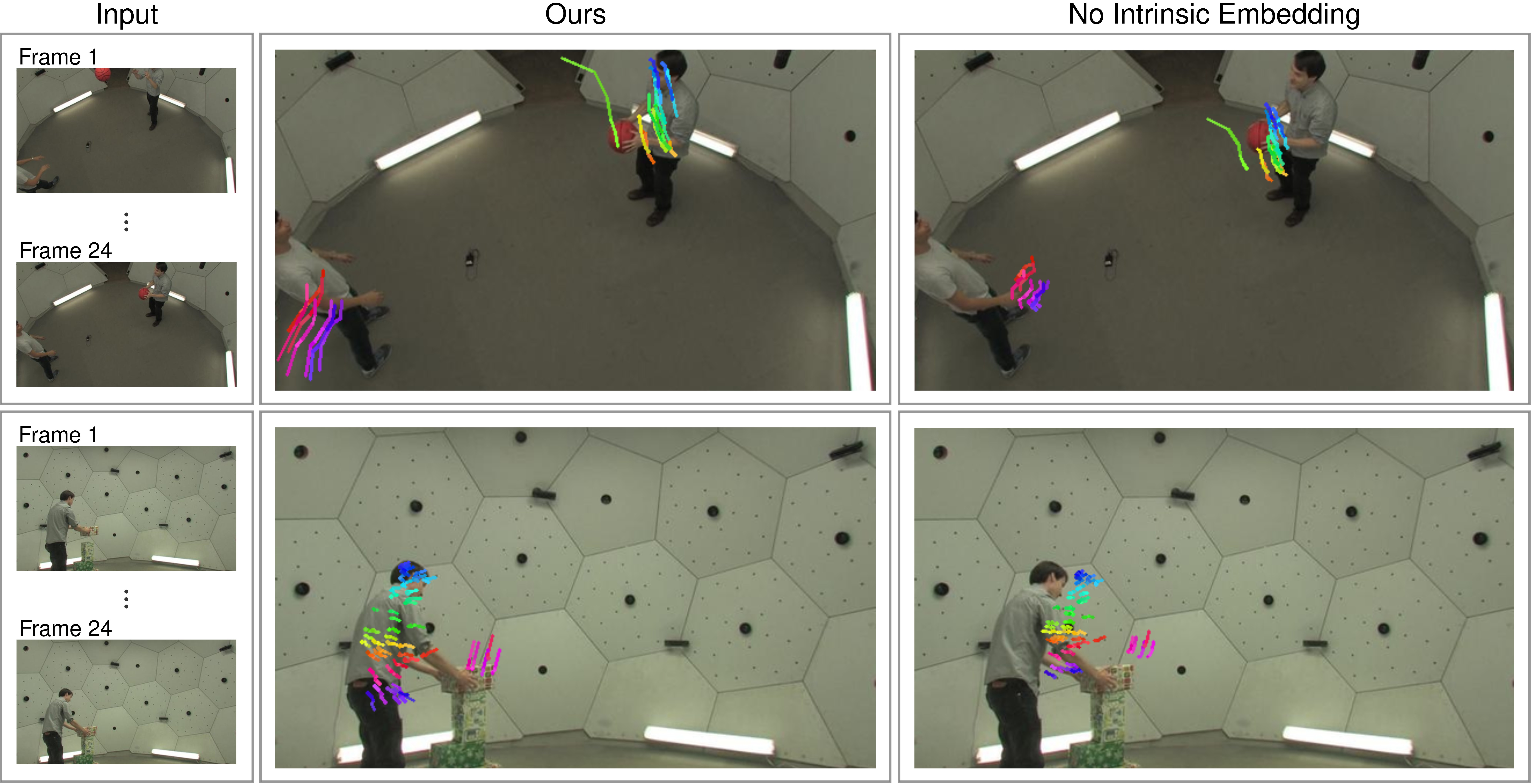}
    \caption{\textbf{Qualitative ablation study on the Panoptic Studio dataset.} The 3D trajectories of scene points are projected onto the last frame. Without the intrinsic embedding, the model can identify the direction of motion but struggles to accurately place it or gauge its magnitude, leading to significant errors in point trajectory estimation. 
    }
    \label{fig:qual}
\end{figure*}

\myparagraph{Ablation Studies.}
Table~\ref{tab:abl_intrinsic} evaluates the roles of the intrinsic embedding and random center-crop augmentation on the Panoptic Studio dataset. 
\begin{itemize} 
    \item \textbf{No intrinsic embedding:} Without the intrinsic embedding, the model struggles to deal with the inherent scale ambiguity in unposed 3D reconstruction and dynamic objects (see Figure~\ref{fig:qual}).  
    \item \textbf{No center-crop augmentation:} Including center-crop augmentation improves both tracking metrics, indicating that training on diverse camera intrinsics is essential for the model to leverage the intrinsic embedding when predicting point motion. 
    \item In conclusion, the best results are obtained when both are used, indicating that the intrinsic embedding, combined with the center crop augmentation, helps the model generalize to out-of-distribution cameras.
\end{itemize}

\begin{table}[t]
    \centering
    \caption{\textbf{Ablation study} of intrinsic embedding and center-crop augmentation on Panoptic Studio dataset. Our experiments demonstrate that the intrinsic embedding is crucial for overcoming the inherent scale ambiguity present in monocular video and achieving generalizable point tracking.
    }
    \label{tab:abl_intrinsic}
    \vspace{-2mm}
    \setlength{\tabcolsep}{6pt}
    \renewcommand{\arraystretch}{1.2}
    \resizebox{0.88\linewidth}{!}{%
    \begin{tabular}{l|cc} %
    \toprule
    \textbf{Variants} & \textbf{APD} ($\uparrow$) & \textbf{EPE} ($\downarrow$) \\
    \midrule \midrule
    No intrinsic embedding & 78.09 & 0.1735 \\
    No center-crop augmentation                    & 82.94 & 0.1398 \\
    \midrule
    DePT3R (Ours) & \textbf{89.36} & \textbf{0.1046} \\
    \bottomrule
    \end{tabular}
    }
    \vspace{-2mm}
\end{table}

\myparagraph{Computational Comparison.} We do not compare with SpatialTrackerV2~\cite{xiao2025spatialtrackerv2}, a recent 3D point-tracking method, because its training data and computational requirements are significantly larger than ours. However, we note that it, along with other state-of-the-art point-tracking methods, struggles to scale to large numbers of query points due to substantial memory demands. To highlight this limitation, we evaluate the GPU memory usage of SpatialTrackerV2, VGGT's 2D point tracker, and our DePT3R for varying numbers of query points. We use a 10-frame video, with a resolution of $518 \times 518$ per frame. SpatialTrackerV2 exceeds 48 GB of memory at 40k query points and ultimately fails due to out-of-memory (OOM) errors, and VGGT exhausts GPU memory at just 22.5k query points. In contrast, our method successfully performs dense point tracking, generating over 268k point tracks ($518 \times 518$ pixels), while consuming only 12 GB of memory.

\begin{figure}[t]
    \centering
    \includegraphics[width=0.99\linewidth]{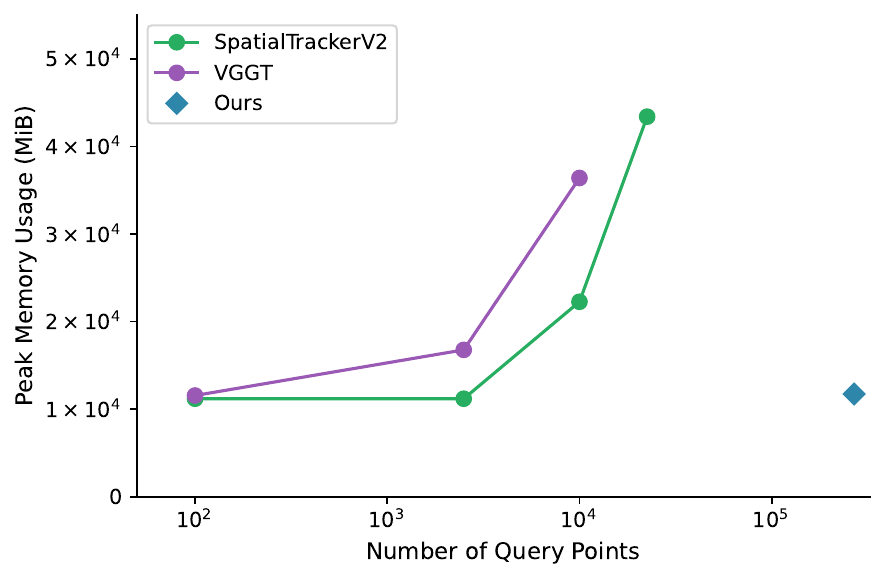}
    \caption{\textbf{GPU memory usage comparison} between SpatialTrackerV2, VGGT and our DePT3R method across varying numbers of query points. SpatialTrackerV2 and VGGT exhibit rapid increases in GPU memory consumption, exceeding the 48 GB limit at just 40k and 22.5k query points, respectively. In contrast, DePT3R efficiently handles 268k query points,  requiring only 12 GB of memory. All experiments were performed on an RTX A6000 GPU (48 GB). Zoom in for better visualization.}
    \label{fig:abl_mem}
\end{figure}

\section{Conclusion and Future Works}
\label{sec:conclusion}

\myparagraph{Summary.} In this paper, we introduce DePT3R, a simple, novel framework for simultaneously performing dense 3D reconstruction and point tracking of dynamic scenes from unposed monocular video. Despite training exclusively on synthetic datasets, our approach achieves strong performance on challenging real-world benchmarks, highlighting the inherent synergy between reconstruction and point-tracking tasks. Unlike prior methods, DePT3R leverages global temporal attention to enhance both accuracy and computational efficiency, enabling dense tracking of all visible points in a single forward pass. Additionally, we effectively leverage intrinsic information to significantly improve reconstruction and motion estimation. 

\vspace{1mm}
\myparagraph{Limitations.} First, while DePT3R uses global attention over multiple frames, it does \textbf{not explicitly model the sequential/causal structure of video} (e.g., temporal recurrence or explicit temporal priors), which may limit robustness under heavy occlusion, motion blur, or very long videos where temporal continuity constraints are beneficial. Second, the current formulation produces motion estimates relative to the \textbf{first-frame coordinate system} and is limited in temporal scope, as described in the paper (i.e., it does not yet provide a general mechanism for \textbf{temporally continuous trajectories} parameterized by arbitrary query times). Third, the approach relies on \textbf{point-tracking annotations} (obtained from synthetic sources in training), which are expensive and difficult to scale in real-world settings; reducing supervision (e.g., via self-/weak-supervised objectives or distillation) remains an important direction. Finally, the current evaluation protocol emphasizes APD/EPE (with global median scaling) and restricts quantitative tracking evaluation to trajectories originating from the first frame; more exhaustive evaluation of dense-any-point tracking and absolute-scale behavior would strengthen conclusions about real deployment.

\vspace{1mm}
\myparagraph{Future Works.} Promising extensions include incorporating \textbf{time-conditioned query mechanisms} (to predict tracks at arbitrary timestamps), explicit \textbf{temporal modeling} (e.g., causal attention or recurrence), and training strategies that reduce dependence on dense trajectory labels while preserving the feed-forward efficiency of the current framework.

\section*{Acknowledgements}
This work was supported by Professor M. Khalid Jawed and, in part, by the US Department of Agriculture (Grant numbers 2024-67021-42528 and 2022-67022-37021).

{
    \small
    \bibliographystyle{ieeenat_fullname}
    \bibliography{main}
}

\end{document}